\documentclass[runningheads]{llncs}
\usepackage{graphicx}

\usepackage{tikz}
\usepackage{comment}
\usepackage{amsmath,amssymb}
\usepackage{color}

\usepackage[accsupp]{axessibility}

\usepackage{hyperref}
\usepackage{threeparttable}

\begin{document}

\pagestyle{headings}
\mainmatter

\title{Unsupervised Difference Learning for Noisy Rigid Image Alignment}

\titlerunning{Preprint}

\author{Yu-Xuan Chen\inst{1} \and
  Dagan Feng\inst{2} \and
  Hong-Bin Shen\inst{1*}}

\authorrunning{Preprint}

\institute{Institute of Image Processing and Pattern Recognition, Shanghai Jiao Tong University, and Key Laboratory of System Control and Information Processing, Ministry of Education of China, Shanghai, 200240, China
  \\ \email{\{yxchen,hbshen\}@sjtu.edu.cn}
  \and School of Computer Science, University of Sydney, Sydney, 2006, Australia\\
  \email{dagan.feng@sydney.edu.au}}

\maketitle

\begin{abstract}
  Rigid image alignment is a fundamental task in computer vision, while the traditional algorithms are either too sensitive to noise or time-consuming. Recent unsupervised image alignment methods developed based on spatial transformer networks show an improved performance on clean images but will not achieve satisfactory performance on noisy images due to its heavy reliance on pixel value comparations. To handle such challenging applications, we report a new {\it u}nsupervised {\it d}ifference {\it l}earning (UDL) strategy and apply it to rigid image alignment. UDL exploits the quantitative properties of regression tasks and converts the original unsupervised problem to pseudo supervised problem. Under the new UDL-based image alignment pipeline, rotation can be accurately estimated on both clean and noisy images and translations can then be easily solved. Experimental results on both nature and cryo-EM images demonstrate the efficacy of our UDL-based unsupervised rigid image alignment method.
  \keywords{Rigid Image Alignment, Unsupervised Learning, Difference Learning, Regression}
\end{abstract}

\section{Introduction}\label{sec:1}

Image alignment which helps establishing pixel-to-pixel correspondence between two images is a fundamental task in computer vision and can be widely applied to tasks like image clustering~\cite{ref001}, 3D image reconstruction~\cite{ref002}, semantic segmentation~\cite{ref003}, and etc. Finding the global optimal parameters of transformation model is an essential and challenging part of image alignment. Traditionally, image alignment was performed in non-trainable approaches in which either hand-crafted features or pixel space were used to find the global optimal parameters of transformation model. In recent years, due to the fast development of deep neural networks (DNN), deep end-to-end image alignment methods have been proposed for both supervised image alignment~\cite{ref004} and unsupervised image alignment~\cite{ref005}. Compared to traditional methods, deep end-to-end image alignment could process a large amount of data in a short time and improve the performance of image alignment by better feature learning on certain datasets.

In supervised image alignment, transformation parameters are directly regressed using DNN with ground truth (GT) transformation parameters required for learning. Although supervised learning achieves promising results, GT labeling is prohibitively expensive and time-consuming in real world applications, which would greatly limit its practical use scope. In \cite{ref005}, end-to-end unsupervised image alignment is achieved by combing DNN with spatial transformer networks (STN)~\cite{ref006}, where DNN is used to regress transformation parameters and STN is used to create a pixel-to-pixel loss function between source image and target image. Though such unsupervised strategy achieves promising results on high quality images, it is not robust to noise. Several methods~\cite{ref007,ref008} are proposed to improve~\cite{ref005} and make it applicable to wider situations by either comparing features instead of input images or designing more complicated loss functions. However, since the loss function of STN-based unsupervised learning is essentially relying on the comparations of pixel values which will be dominated by noise in low signal-to-noise-ratio (SNR) cases, the STN-based unsupervised strategy is still not very robust to noisy image alignment applications.

To tackle image alignment in noisy situations, we propose a new unsupervised difference learning strategy UDL and design unsupervised rigid image alignment algorithm based on it in conjunction with network architectures. Our UDL-based image alignment method is designed for rigid image transformation containing only in-plane rotation and translations, which is essential in image alignment and common in certain cases like medical images registration~\cite{ref009} and cryo-electron microscopy (cryo-EM) single particle image alignment~\cite{ref002}. Unlike STN-based unsupervised learning methods which rely on pixel values to guide network learning, UDL strategy use the difference value between network outputs to create pseudo labels, so that unsupervised learning can be achieved using the same architecture as supervised learning. Utilizing such pseudo labels to train DNN, our unsupervised learning strategy can be generalized to multiple challenging situations just like the supervised learning pipelines. In particular, UDL strategy can provide an unsupervised learning solution to noisy rigid image alignment.

We demonstrate the simplicity and effectiveness of UDL strategy by applying it to synthetic MS-COCO dataset~\cite{ref010}, synthetic cryo-EM single particle dataset and real-world cryo-EM single particle dataset. Experimental results show that we achieve competitive results as STN-based unsupervised learning methods on clean images, better results than supervised learning method on noisy images and better results than traditional Fourier-based method on cryo-EM images, suggesting the UDL-based rigid image alignment method works for both clean and noisy images. To summarize, our main contributions are:
\begin{enumerate}
  \item[-] We propose a new unsupervised difference learning strategy UDL, which converts unsupervised learning to supervised learning utilizing the difference of network outputs to generate pseudo labels.
  \item[-] We apply UDL to noisy rigid image alignment and demonstrate the performance of our UDL-based rigid image alignment method on both nature and cryo-EM datasets.
  \item[-] We suggest that UDL can also be extended to other regression tasks.
\end{enumerate}

\section{Related Work}\label{sec:2}

\subsection{Traditional Image Alignment}\label{sec:2.1}

Traditional image alignment methods can be roughly divided into two categories~\cite{ref011}, feature-based methods and area-based methods. Featured-based methods calculate transformation parameters by detecting and matching local features. After matched feature pairs obtained by methods such as SIFT~\cite{ref012} and ORB~\cite{ref013}, transformation parameters can be estimated by algorithms such as RANSAC~\cite{ref014}. Area-based methods directly use image similarity like cross-correlation (CC) and sequential similarity detection algorithm (SSDA)~\cite{ref015} to find the best matching position, and the progress could be speeded up by Fourier transform~\cite{ref016}. Although featured-based methods usually outperform area-based methods, their performance heavily depends on the quality of hand-crafted features and input images. In noisy images such as cryo-EM images where SNR is typically lower than 0.1, local feature detection would be dominated by noise and thus the performance of featured-based methods would degrade. Though area-based methods would be more robust in such noisy cases, they are usually of intensive computational cost, and their image similarity calculations are also subject to noise.

In the typical cryo-EM single particle noisy image alignment, algorithms in software such as XMIPP~\cite{ref017} and SPIDER~\cite{ref018} use iterative strategy to reduce alignment error caused by noise. Though iterative strategies help on noisy situations, it is generally computationally expensive. EMAF~\cite{ref019} uses hand-crafted features on Fourier space to directly estimate transformation, but its practical use is limited due to its Gaussian noise assumption.

\subsection{Supervised Image Alignment}\label{sec:2.2}

Based on the powerful feature learning ability of convolutional neural networks (CNN), deep learning has been applied to image alignment firstly in \cite{ref004}, which builds a VGG like architecture~\cite{ref020} and uses four corner displacements to describe transformation model. This method achieves better results than traditional image alignment methods on homography estimation. To improve the architecture of \cite{ref004}, feature correlation and matching layer are introduced in \cite{ref021} to help learning correspondence between input images so that more accurate transformation parameters can be estimated. Zeng {\it et al.}~\cite{ref022} uses a U-Net like architecture~\cite{ref023} to estimate the pixel-to-pixel bijection and then transformation parameters are calculated using direct linear transformation (DLT)~\cite{ref024} and RANSAC. Although supervised methods achieve promising results on both clean and noisy images, lack of the ground truth labels limits their use in the real world applications.

\subsection{Unsupervised Image Alignment}\label{sec:2.3}

Based on differentiable image warping proposed in STN, unsupervised image alignment proposed in \cite{ref005} adopts four corner displacements to describe homography transformation model and uses STN to warp source image so that the pixel-to-pixel density error can be calculated and minimized. Such unsupervised strategy achieve comparable results as \cite{ref004} on synthetic MS-COCO dataset, but will face challenges when dealing with large lamination change or noisy images. To make \cite{ref005} more robust, Zhang {\it et al.}~\cite{ref007} applies pixel-to-pixel minimization to features instead of images and maximizes feature difference of two input images to avoid trivial solutions. However, this method is designed for small scale alignment and can be difficult on big changes. Koguciuk {\it et al.}~\cite{ref008} introduces bidirectional implicit homography estimation (biHomE) loss to decouple transformation parameters estimation from the representation learning and make the approach more robust. With the advancement of Transformers~\cite{ref025} in natural language processing (NLP), the Vision Transformer (ViT)~\cite{ref026} attracts much attention in computer vision field. In \cite{ref027}, ViT is applied to medical image alignment for deformation field estimation and then STN is used to minimize pixel-to-pixel density error between warped source image and target image. Nevertheless, since the loss function of STN-based unsupervised learning essentially relies on pixel values, its performance on noisy situations has much space to improve.

\section{Methods}\label{sec:3}

\subsection{Unsupervised Difference Learning Strategy UDL}\label{sec:3.1}

In general, when GT labels are available, supervised learning is more robust than unsupervised learning and can adapt for more situations. In supervised learning, network outputs are compared to GT labels to guide loss backpropagation. However, since GT labels are not always available in real world applications, supervised problems often become unsupervised problems. On such unsupervised cases, we find that differences between network outputs are interpretable for some regression tasks, which means output differences can be used as pseudo labels to guide learning. Based on this intuitive idea, we propose an unsupervised difference learning strategy UDL in this paper for certain regression tasks including rigid image alignment.

Denote two inputs to network as $x_1$ and $x_2$, their corresponding outputs as $y_1$ and $y_2$, their corresponding GT labels as $\hat{y_1}$ and $\hat{y_2}$, and the network itself as $F_{\theta}$, where $\theta$ denotes network parameters; then the forward progress of network can be represented as $y_1=F_{\theta}(x_1)$ and $y_2=F_{\theta}(x_2)$.

If there exist function $G$ and function $H$ which satisfy $G(\hat{y_1},\hat{y_2})=H(x_1,x_2)$, then pseudo labels can be created for loss function and the corresponding supervised training with pseudo labels can be carried on even though the original GT labels are unknown; that is:
\begin{align}\label{eq:1}
  \min_{\theta} ||G(F_{\theta}(x_1),F_{\theta}(x_2))-H(x_1,x_2)||
\end{align}

In the most ideal case, Eq.~\ref{eq:1} is equal to supervised training with GTs so that correct outputs can be obtained from well-trained network by pseudo labels. In such case, Eq.~\ref{eq:1} would be equal to:
\begin{align}\label{eq:2}
  \min_{\theta} ||F_{\theta}(x_1)-\hat{y_1}||
\end{align}

However, although the convergence of Eq.~\ref{eq:2} would always lead to the convergence of Eq.~\ref{eq:1}, the inverse progress does not always hold. Ignoring the training progress and assuming network will always converge once GT labels are given, from the perspective of equation solving, there is one unknown variable $y_1$ (i.e. $F_{\theta}(x_1)$) to solve in supervised training problem and one equation $y_1=\hat{y_1}$ given in format of loss function. Whereas in unsupervised training problem, no equation is provided but there is one unknown variable $y_1$ to solve. In supervised training with pseudo labels, there are two unknown variables $y_1$ and $y_2$ to solve but only one equation Eq.~\ref{eq:1} is provided, which leads to an indeterminate equation whose solution space consists of infinite solutions. By designing Eq.~\ref{eq:1}, the original unsolvable problem is converted to a solvable problem with infinite possible solutions, but only one of them is the correct solution of target problem Eq.~\ref{eq:2}.

In this way, the theoretical convergence of the designed pseudo supervised training could not be guaranteed due to the training progress could converge on any solution in solution space instead of the desired one. However, since the solutions are heavily subject to the form of Eq.~\ref{eq:1} and thus subject to the formulation of function G and H, by properly designing these two functions, solutions in solution space of Eq.~\ref{eq:2} could be obtained in a trackable form, which then leads to the convergence of pseudo supervised training.

Based on this idea, function G is designed as the difference value between $F_{\theta}(x_1)$ and $F_{\theta}(x_2)$. For instance, difference value could be $F_{\theta}(x_1)-F_{\theta}(x_2)$ in simple numeric tasks, transformation between two pairs of images in image alignment tasks and bounding box movements in object recognition tasks. Such design of function G would make Eq.~\ref{eq:2} be in similar form of the derivative of Eq.~\ref{eq:1}, and leads to simple form of solutions (i.e. the desired solution plus a fixed bias). Then the training convergence can be kept at the most degree. However, even if function $G$ is designed in such simple form, function $H$ is hard to formulate for any two sampled inputs. Therefore, to better formulate the relationship between two inputs, we generate input $x_2$ from input $x_1$ by manually designing disturbance to create output differences.

Denote $x_2$ as $D_{\tau}(x_1)$ where $D$ represents the manually designed disturbance function and $\tau$ represents the scale of output difference caused by disturbance; then Eq.~\ref{eq:1} becomes:
\begin{align}\label{eq:3}
  \min_{\theta} ||G(F_{\theta}(x_1),F_{\theta}(D_{\tau}(x_1)))-H(x_1,D_{\tau}(x_1))||
\end{align}
In Eq.~\ref{eq:3}, function $G$ should calculate output difference as close as possible to the manually designed difference $\tau$ and thus leads to convergence of pseudo supervised training. Then function $G$ and $H$ are:
\begin{align}\label{eq:4,5}
  G: F_{\theta}(x_1), F_{\theta}(D_{\tau}(x_1) \rightarrow \tau \\
  H: x_1, D_{\tau}(x_1) \rightarrow \tau
\end{align}

As mentioned before, even though the convergence of pseudo supervised training can be achieved, network output $y_1$ could be different from GT label $\hat{y_1}$ due to infinite solutions of indeterminate equation. Denote the difference function as $T(F_{\theta}(x_1 ))-T(F_{\theta}(x_2 ))$, where $T$ could be simply identity function or in other formats to convert difference function to subtractive format; then network outputs would be $T(F_{\theta}(x))+C$. Such progress is just like recovering a function from its derivative. If disturbance is well-designed, $C$ would be a constant value and easy to determine on any input in dataset with only one GT required. Though detailed design of disturbance depends on specific tasks, the core idea is to minimize shared elements in original inputs and disturbed inputs so that $C$ is less likely to generate from the shared elements of network inputs. The specific disturbance of UDL for rigid image alignment would be discussed in the following section.

\subsection{Unsupervised Image Alignment Based on UDL}\label{sec:3.2}

\begin{figure}
  \centering
  \includegraphics[height=6cm]{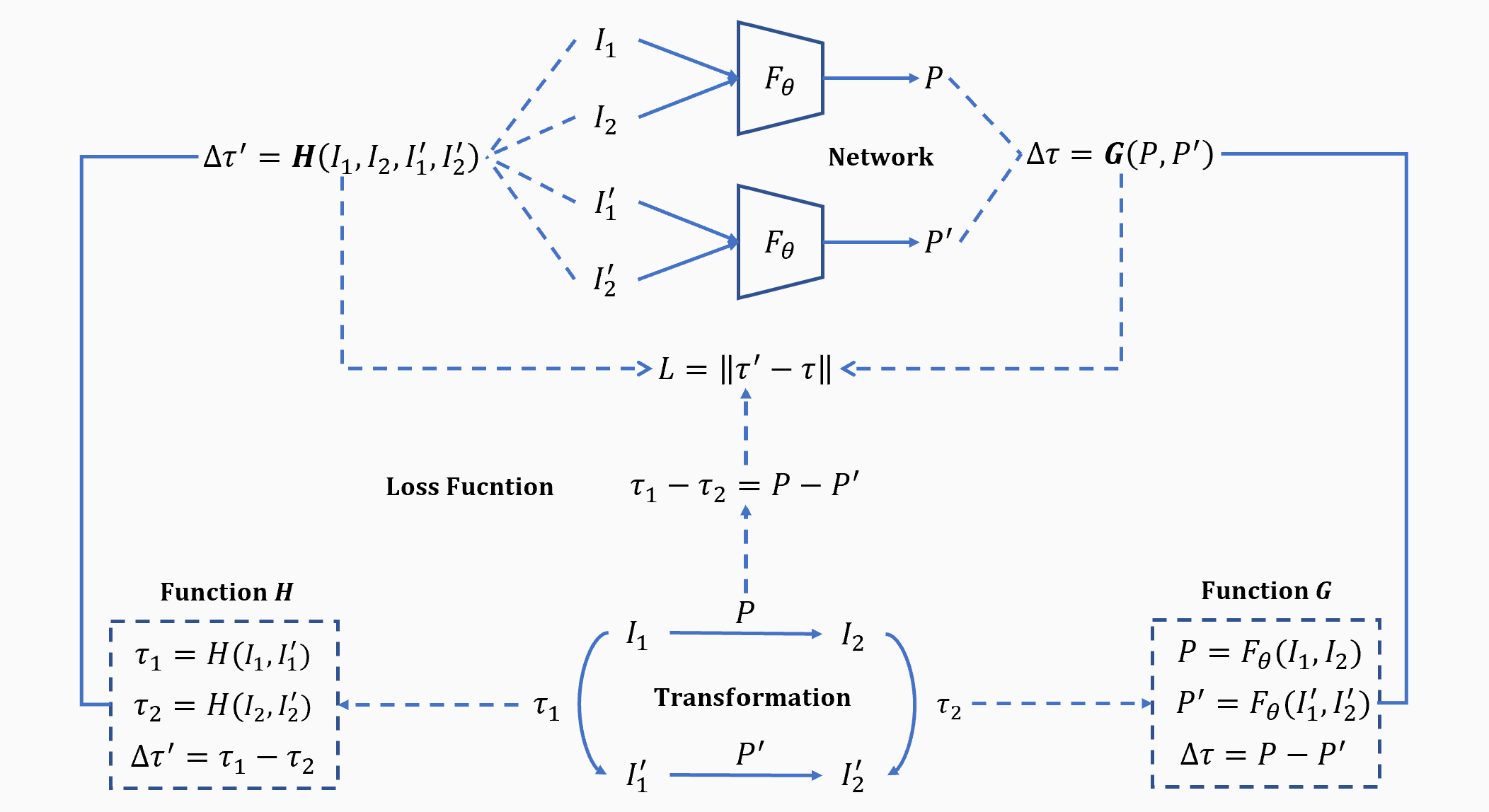}
  \caption{Unsupervised difference learning strategy UDL for rigid image alignment using symbols defined in Sect.~\ref{sec:3.1}. In functions $G$ and $H$, the minus sign represents transformation parameters that one image needs to transform into another image.}
  \label{fig:1}
\end{figure}

As mentioned before, we focus on rigid image alignment which contains only in-plane rotation and translations. Network input is a pair of images and network outputs are their relative transformation parameters which could be in formats like two corner displacements (which are enough to cover rigid transformation), transformation matrix or simply one rotation angle and two translations. Based on the proposed unsupervised difference learning strategy UDL in Sect.~\ref{sec:3.1}, the unsupervised rigid image alignment pipeline is shown in Fig.~\ref{fig:1}.

As shown in Fig.~\ref{fig:1}, input image $I_1$ is transformed into disturbed image $I'_1$ by parameter $\tau_1$; input image $I_2$ is transformed into disturbed image $I'_2$ by parameter $\tau_2$; network outputs for image pairs $(I_1,I_2)$ and $(I'_1,I'_2)$ are correspondingly $P$ and $P'$. Disturbances are designed as rotation and translations on source images.

For any output format, if quantitative transformation relationship is going to establish between two images, the output must be converted to transformation matrix which projects every pixel in source image to target image. In Fig.~\ref{fig:1}, if image $I_1$ is likely to transform into image $I'_2$, there are two paths, i.e. $I_1 \rightarrow I_2 \rightarrow I'_2$ and $I_1 \rightarrow I'_1 \rightarrow I'_2$. Denote transformation matrices for $\tau_1$, $\tau_2$, $P$ and $P'$ are correspondingly $M_1$, $M_2$, $M$ and $M'$, and use image  to represent all its pixels; then quantitative transformation relationship can be established as:
\begin{align}\label{eq:6}
  I'_2=M_2MI_1, \hspace{1cm} I'_2=M'M_1I_1
\end{align}
which leads to:
\begin{align}\label{eq:7}
  M_2M=M'M_1
\end{align}
Then the loss function can be designed as:
\begin{align}\label{eq:8}
  L_M=||M_2M-M'M_1||
\end{align}

For rigid transformation, transformation matrix is of certain format:
\begin{align}\label{eq:9}
  M=\begin{bmatrix}
    \cos\alpha  & \sin\alpha & \Delta x \\
    -\sin\alpha & \cos\alpha & \Delta y \\
    0           & 0          & 1
  \end{bmatrix}
\end{align}
where $\alpha$ is rotation angle, $\Delta x$ is translation in $X$ axis and $\Delta y$ is translation in $Y$ axis. By matrix multiplication, Eq.~\ref{eq:8} leads to an indeterminate equation system of 4 independent equations and 6 unknown variables. If output formats like corner displacements are used, $\sin\alpha$ and $\cos\alpha$ would be considered as two parameters and even more unknown variables would be produced. Such a severe imbalance of equations and variables makes Eq.~\ref{eq:8} hard to converge on target solution. The form of multivariate linear equations generated from matrix multiplication also implies differences between  solutions in solution space of indeterminate equation system cannot be simply described in a subtractive format like $T(F_{\theta}(x))+C$ in Sect.~\ref{sec:3.1}.

Based on the above considerations, we adopt a strategy of separating rotation and translation, in which we use UDL to estimate only the rotation and then apply Fourier alignment to re-rotated images to estimate translations. Since only rotation is estimated in UDL, network can simply output rotation angle between two images. Then Eq.~\ref{eq:7} and Eq.~\ref{eq:8} become:
\begin{align}
  \label{eq:10}
  \alpha_2+\alpha=\alpha'+\alpha_1 \Rightarrow \alpha-\alpha'=\alpha_1-\alpha_2 \\
  \label{eq:11}
  L_{\alpha}=||(\alpha-\alpha')-(\alpha_1-\alpha_2)||
\end{align}
where $\alpha_1$, $\alpha_2$, $\alpha$ and $\alpha'$ are the corresponding rotation angles of $M_1$, $M_2$, $M$ and $M'$ and also the correspondingly values of $P_1$, $P_2$, $P$ and $P'$. Then function $G$ would be $G=P-P'=(\alpha-\alpha')$ and function H would be $H=H_{\alpha_1}(I_1,I'_1 )-H_{\alpha_2}(I_2,I'_2 )=\alpha_1-\alpha_2$, where $I'_1$, $I'_2$ are generated from $I_1$, $I_2$ by rotation angle $\alpha_1$, $\alpha_2$ and random translations in a pre-defined range.

Since rotation angle is an infinite continuous periodic signal and we expect its value locates in period $[0,r]$($[0^\circ,360^\circ]$  in implementation), Eq.~\ref{eq:11} is modified as follows by considering the signal period and range penalty:
\begin{align}\label{eq:12}
  L=L_r((\alpha-\alpha')\%r,(\alpha_1-\alpha_2)\%r)+P_r(\alpha)+P_r(\alpha')
\end{align}
with
\begin{align}
  \label{eq:13}
  L_r(a,b)=\left\{\begin{array}{l}
    \min(\min(a-r,b),\min(a,b)), a \geq r/2 \\
    \min(\min(a+r,b),\min(a,b)), a < r/2
  \end{array}\right. \\
  \label{eq:14}
  P_r(a)=\max(0,-a)+\max(0,r-a)
\end{align}
where $L_r(a,b)$ is a function considering period affection with the premise that $a$,$b$ are both in range $[0,r]$, and $P_r(a)$ is penalty function to force network outputs being in range $[0,r]$.

In our implementation, both two input images are disturbed to reduce shared elements in original inputs and disturbed inputs. After network training, network outputs $\alpha+C$ where $C$ is a constant value and can be determined on any pair of input images. After rotation angle estimation, translations can be simply and quickly estimated by Fourier correlation~\cite{ref016}, which is robust to noise for translation only image alignment.

\subsection{Network Architecture}\label{Sec:3.3}

\begin{figure}
  \centering
  \includegraphics[height=5cm]{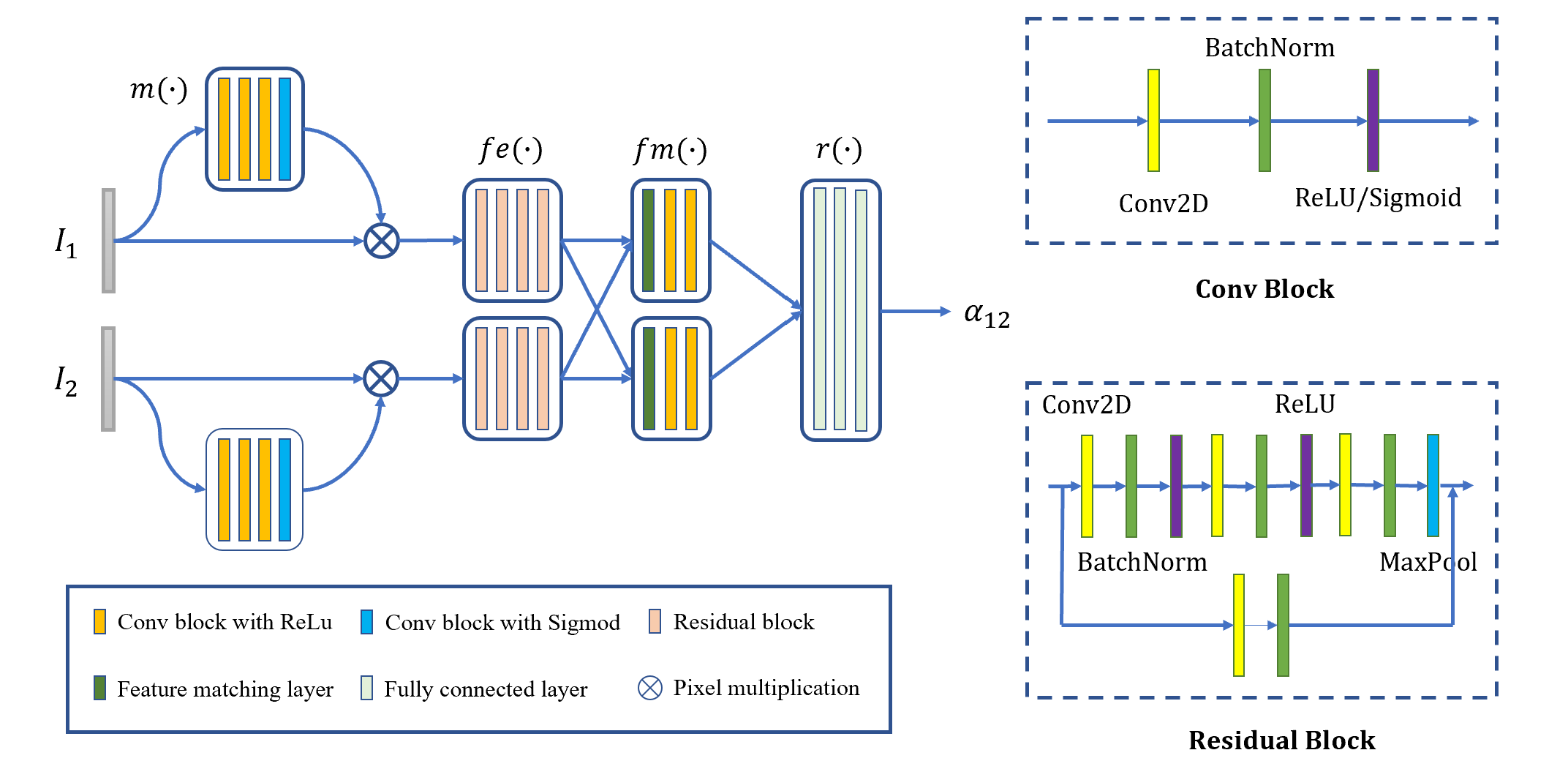}
  \caption{The overall architecture of our rigid image alignment network using the unsupervised difference learning strategy UDL.}
  \label{fig:2}
\end{figure}

Our network architecture is built upon CNN and the overall architecture is shown in Fig.~\ref{fig:2}. The network takes in two images $I_1$, $I_2$ as input and outputs their relative rotation angle $\alpha_{12}$. The entire network architecture consists of mask prediction module $m(\cdot)$, feature extraction module $fe(\cdot)$, feature matching module $fm(\cdot)$ and regression module $r(\cdot)$. All modules except for regression module are replicated in a siamese configuration so that two input images pass through two identical networks with shared parameters.

{\it Mask prediction module}. We adopt mask prediction module proposed in \cite{ref007} to automatically learn the positions of significant areas in input images. Especially in cryo-EM single particle images, particles are located near the centers of images and surrounded by meaningless background, where transformation relationship only holds between particles instead of images. Mask prediction module would help the network focus on particles and learn features containing richer transformation information. Mask prediction module consists of four convolutional blocks with the last one using Sigmoid activate function instead of ReLU. Mask prediction module takes in image $I \in R^{1\times h \times w}$ and outputs mask $M \in R^{1\times h \times w}$.

{\it Feature extraction module}. Feature extraction module consists of four residual blocks~\cite{ref028} to learn deep features for image alignment. Unlike architecture in \cite{ref004,ref007,ref008} where an entire ResNet or VGG is used for feature extraction, we use several residual blocks cropped at middle stage to extract features so that there exists enough spatial information for the following feature matching module. Feature extraction takes in image with mask $M \times I \in R^{1\times h \times w}$ and outputs feature $F \in R^{256\times (h/16) \times (w/16)}$.

{\it Feature matching module}. We adopt feature matching layer proposed in \cite{ref021} to utilize pairwise descriptor similarities and their spatial locations to estimate transformation parameters. Feature matching layer is essentially a normalized cross-correlation function in which all channels of one feature location is compared to all locations of another feature to generate correlation map. The feature matching layer takes in two features $F \in R^{c\times h \times w}$ as input and outputs their correlation map $C \in R^{(h\times w)\times h \times w}$. Siamese structure proposed in \cite{ref029} is adopted to explore more spatial correlation information and two following convolutional blocks after the feature matching module are used to help further enrich spatial features.

{\it Regression module}. After feature matching, two correlation maps are flattened into one dimensional array, concatenated together and feed into fully connected layers. Regression module consists of three fully connected layers and outputs the relative rotation angle between $I_2$ and $I_1$.

\section{Experimental Results}\label{sec:4}

\subsection{Datasets and Implementation Details}\label{sec:4.1}

\begin{figure}
  \centering
  \includegraphics[height=5.5cm]{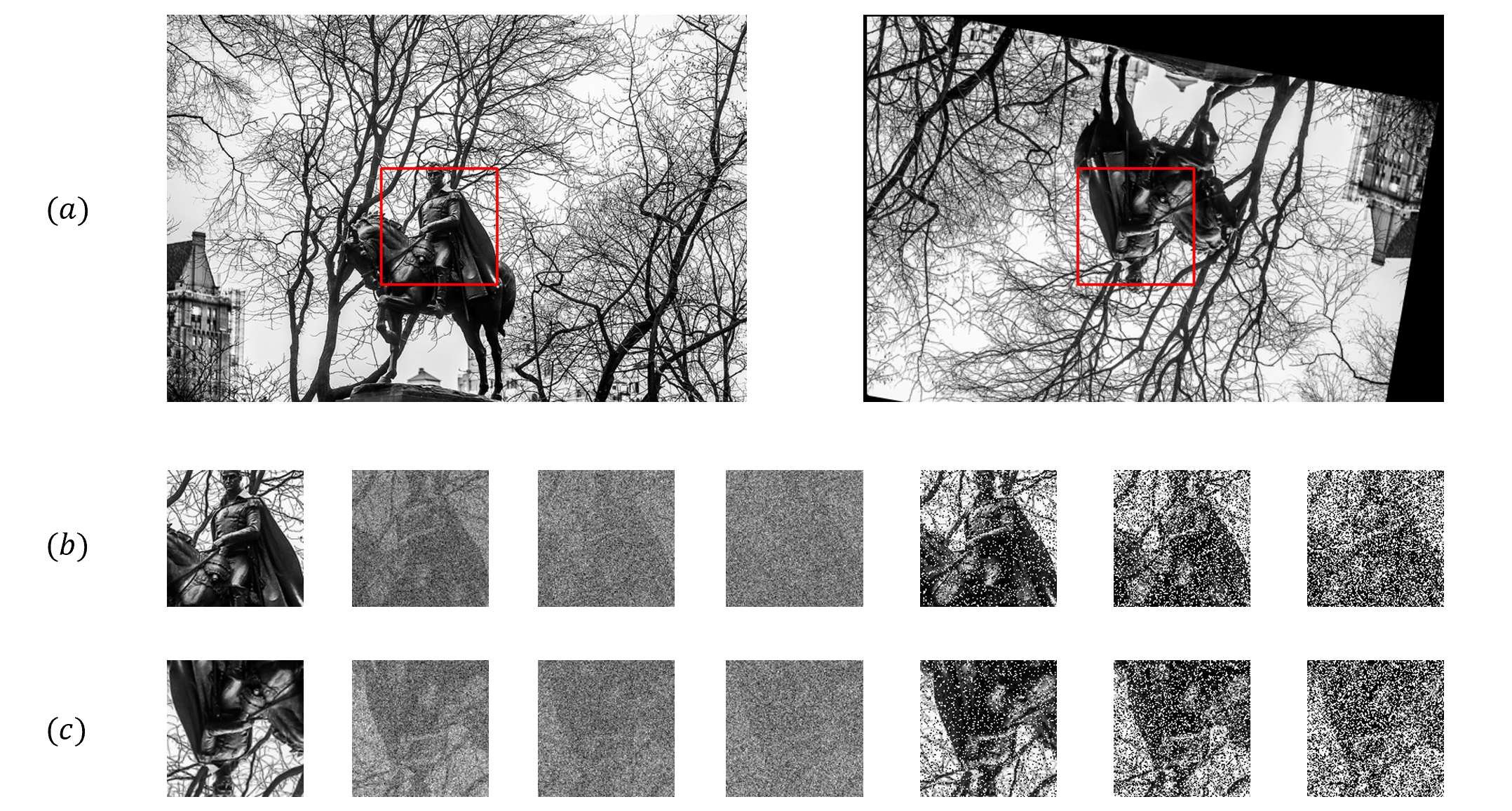}
  \caption{Process for generating synthetic MS-COCO dataset. (a) apply rotation and translations to source image and extract patches on two images at the same position. (b) patches from source image. (c)  patches from transformed image. From left to right in (b) and (c): clean, Gaussian SNR 0.5, Gaussian SNR 0.1, Gaussian SNR 0.05, salt-and-pepper proportion 0.1, salt-and-pepper proportion 0.2, salt-and-pepper proportion 0.3. }
  \label{fig:3}
\end{figure}

Since collecting data for image alignment with ground truth is hard, based on similar pipeline of benchmark dataset generation in \cite{ref004,ref007,ref008}, we generate synthetic MS-COCO dataset for rigid alignment evaluation. Each image in MS-COCO dataset is rotated with random rotation angle ranging in $[0^\circ,360^\circ]$ and shifted with random translation pixels ranging in $[0,10]$ for both horizontal and vertical directions (i.e. left or right, up or down). Then two image patches of size $128\times 128$ are extracted at the same position of both original image and transformed image to generate image alignment patch pairs. If noise is required, Gaussian noise is added to each image patch to make its SNR being certain value (0.5, 0.1 and 0.05 in implementation) or salt-and-pepper noise is added to each image patch with certain proportion for both salt noise and pepper noise (0.1, 0.2, 0.3 in implementation). There are about 115k image pairs for training and 5k image pairs for evaluation. Image patch is converted to grayscale as \cite{ref004,ref005,ref007,ref008}. The data preparation process is shown in Fig.~\ref{fig:3}.

\begin{figure}
  \centering
  \includegraphics[height=4cm]{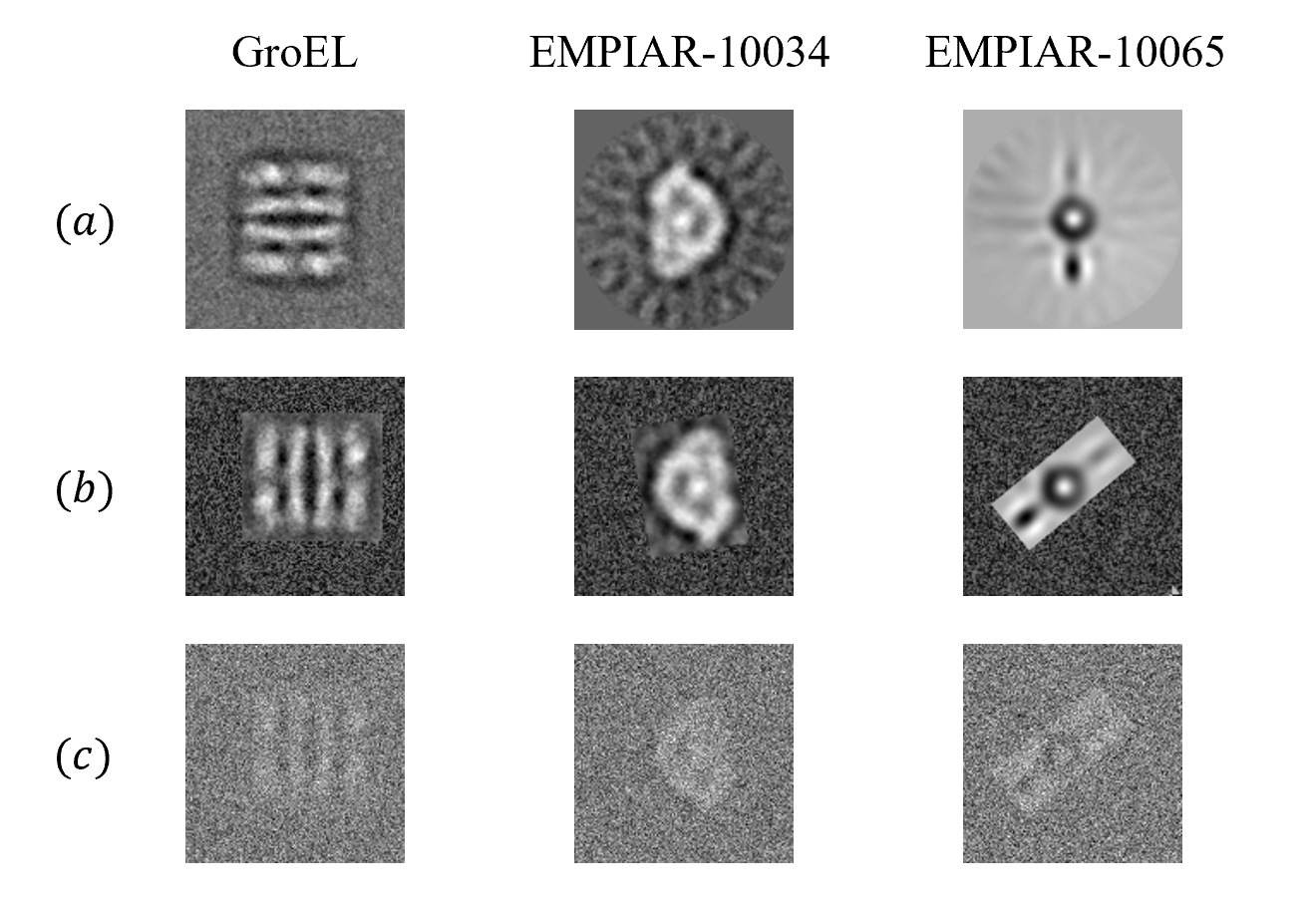}
  \caption{From left to right in the figure in each line: GroEL, EMPIAR-10034, EMPIAR-10065. (a) three cluster centers as source image for synthetic cryo-EM datasets. (b) clean images generated with random background. (c) Gaussian noisy images generated from clean images of SNR 0.1.}
  \label{fig:4}
\end{figure}

In cryo-EM datasets, transformation relationship only holds on part of the entire image (i.e. particles) and SNR of single particle images is typically lower than 0.1, making the alignment task challenging. Although the format of real-world cryo-EM noise is complicated, as depicted in \cite{ref025}, it is generally considered as Gaussian distribution. Such assumption is widely used in the field~\cite{ref030,ref031,ref032} and also adopted in this paper. As shown in Fig.~\ref{fig:4}, we use datasets from EMPIAR~\cite{ref033} and real-world dataset GroEL~\cite{ref034} to generate three cluster centers as source images by existing clustering method CL2D~\cite{ref035}. Then particle in each source image is extracted to generate synthetic datasets of 50k images of size $128\times 128$ with random surrounded background, random rotation angle ranging in $[0^\circ,360^\circ]$, random translation pixels ranging in $[0,10]$, and Gaussian noise to make SNR of each image being 0.1. In addition to synthetic cryo-EM datasets with GT labels to quantify performance of alignment methods, we also use real-world cryo-EM dataset GroEL~\cite{ref034} to validate our method. Since no GT label is available for real-world cryo-EM datasets, we align every image to a given reference image and calculate similarity between aligned image and reference image as a similarity score, as better alignment method will yield higher similarity.

In implementation, our network is trained with batch size 64 by Adam optimizer~\cite{ref036} with parameters $l_r=5.0\times {10}^{-4}$, $\beta_1=0.9$, $\beta_2=0.999$ and $eps=1.0\times {10}^{-8}$. For every 12k iterations, learning rate $l_r$ is reduced by $20\%$. Convolutional block consists of one 2D convolutional layer of kernel size $3\times 3$, a batch normalization layer~\cite{ref037} and an activation layer ReLU or Sigmoid. Residual block consists of three 2D convolutional layers of kernel size $1\times 1$, $3\times 3$ and $1\times 1$ with the first two followed by batch normalization layer and ReLU layer, and the last one only followed by batch normalization layer. Shortcut consists of a 2D convolutional layer of kernel size $1\times 1$ and a batch normalization layer. All maxpooling layers use down-sampling factor 2. The regression module consists of three fully connected layers of output sizes 2000, 2000 and 1. All methods including STN-based unsupervised method, our UDL-based method and supervised method use the same architecture with only the last fully connected layer replaced in STN-based unsupervised method to output two corner displacements (i.e. 4 parameters). We implement all code in PyTorch~\cite{ref038} and release the source code for public use.

\subsection{Results on Synthetic MS-COCO Dataset}\label{sec:4.2}

\begin{figure}
  \centering
  \includegraphics[height=5cm]{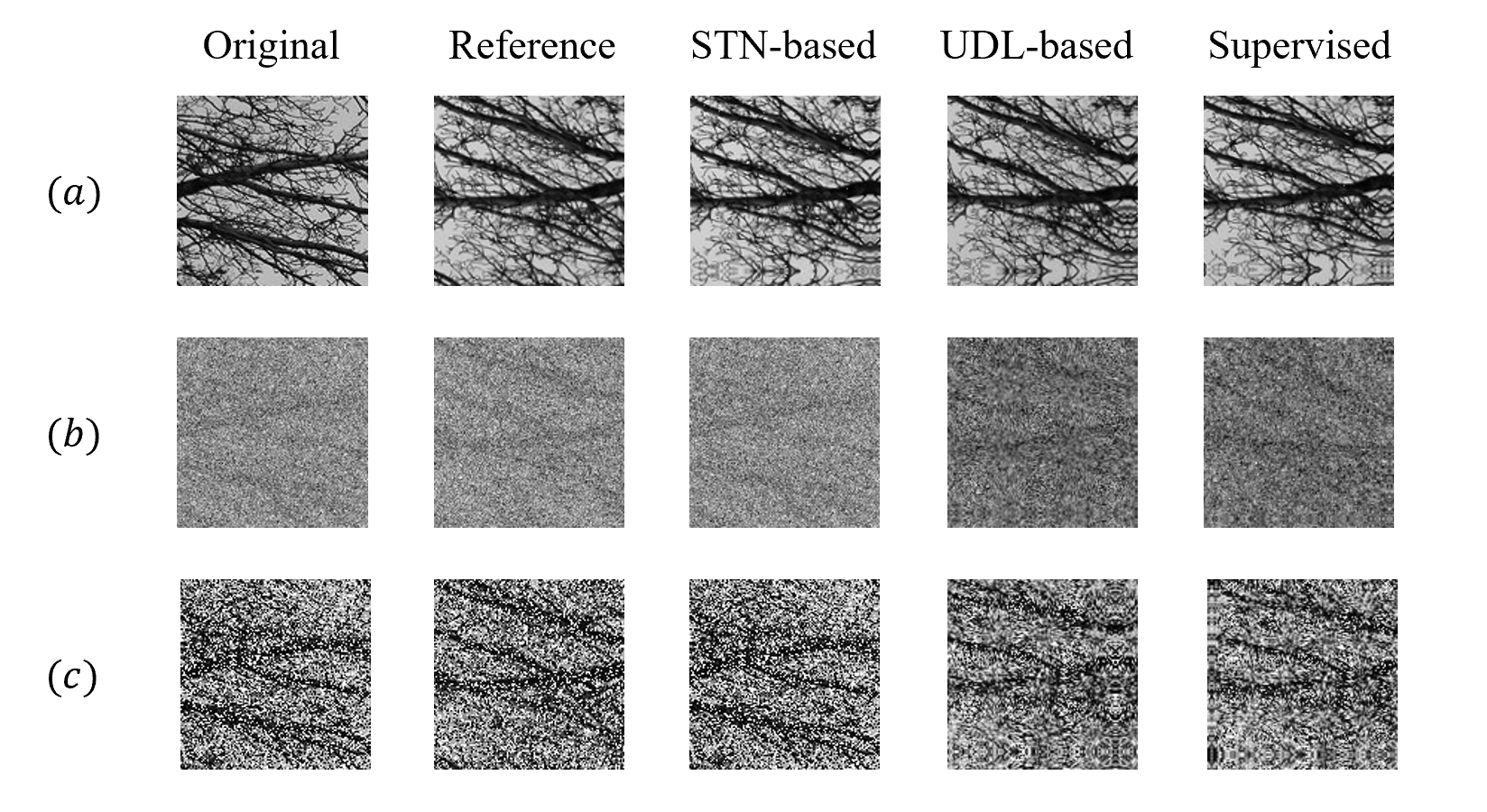}
  \caption{From left to right in the figures in each line: original image, reference image, results of STN-based method, results of our UDL-based method, results of supervised method. (a) Clean dataset. (b) Gaussian SNR 0.1 noisy dataset. (c) Salt-and-pepper proportion 0.2 noisy dataset.}
  \label{fig:5}
\end{figure}

We perform experiments on clean, Gaussian noisy and salt-and-pepper noisy datasets to compare STN-based unsupervised method, our UDL-based unsupervised method and supervised method. All methods use 115k pair of image patches for training and 5k pair of image patches for testing, where training set and test set were pre-separated by MS-COCO dataset and patches of little textures are discarded. As depicted in Sect.~\ref{sec:3.2}, our UDL-based method outputs only the rotation angle. Since translations are determined by traditional Fourier alignment once correct rotation is given, to concentrate on evaluating unsupervised methods, we simply report rotation angle error of each method. Supervised method and our UDL-based method directly output rotation angle and STN-based method outputs two corner displacements which is then converted to rotation angle for comparation. Results of rotation angle error in test set are shown in Table~\ref{tab:1} and sampled output images are shown in Fig.~\ref{fig:5}.

\setlength{\tabcolsep}{4pt}
\begin{table}
  \centering
  \caption{Rotation angle error for synthetic MS-COCO datasets (in degree).}
  \label{tab:1}
  \begin{threeparttable}[b]
    \begin{tabular}{lccc}
      \hline\noalign{\smallskip}
                                     & Supervised & STN-based    & UDL-based  \\
      \noalign{\smallskip}
      \hline
      \noalign{\smallskip}
      Clean                          & 3.87       & {\bf 0.96}   & 2.57       \\
      Gaussian SNR 0.5               & 6.78       & N/A\tnote{1} & {\bf 4.29} \\
      Gaussian SNR 0.1               & 9.23       & N/A\tnote{1} & {\bf 5.69} \\
      Gaussian SNR 0.05              & 12.27      & N/A\tnote{1} & {\bf 6.49} \\
      Salt-and-pepper proportion 0.1 & 5.72       & N/A\tnote{1} & {\bf 3.38} \\
      Salt-and-pepper proportion 0.2 & 5.47       & N/A\tnote{1} & {\bf 4.92} \\
      Salt-and-pepper proportion 0.3 & {\bf 6.10} & N/A\tnote{1} & 12.10      \\
      \hline
    \end{tabular}
    \begin{tablenotes}
      \footnotesize
      \item[1] Not converge in our experiments.
    \end{tablenotes}
  \end{threeparttable}
\end{table}
\setlength{\tabcolsep}{1.4pt}

As shown in Table~\ref{tab:1}, STN-based unsupervised method outperforms the other two methods on clean dataset but does not converge in either Gaussian noisy dataset or salt-and-pepper noisy dataset in our experiments. Its heavy reliance on pixel value comparations greatly improves its accuracy on clean dataset where pixel value comparations are reliable but may also cause its failure on noisy datasets where pixel value comparations are dominated by noise. In such noisy cases, supervised method shows its robustness though its performance degrades on test set due to overfitting. Our proposed UDL-based unsupervised method has comparable or even better performance than supervised method on most datasets, and suffers less from overfitting due to unsupervised training. Such results demonstrate the correctness and effectiveness of our UDL strategy and UDL-based unsupervised rigid image alignment method.

\subsection{Results on Cryo-EM Dataset}\label{sec:4.3}

To better explore the usage of our UDL-based unsupervised rigid image alignment method, we conduct experiments on both synthetic and real-world cryo-EM single particle images to evaluate performance. As depicted in \cite{ref019}, Fourier domain makes it easier to distinguish particle from noise and background in cryo-EM images compared to spatial domain, which also keeps exactly the same rotation relationship as spatial domain. Therefore, though keeping the network architecture and the difference learning strategy UDL unchanged, we replace inputs images to network with their Fourier spectrums to improve the training convergence. As illustrated in Sect.~\ref{sec:4.1}, we use GroEL, EMPIAR-10034, EMPIAR-10065 to generate three synthetic cryo-EM single particle datasets for quantitative comparison. We compare it to traditional Fourier-based method~\cite{ref019} and show results in Table~\ref{tab:2}.
\setlength{\tabcolsep}{4pt}
\begin{table}
  \centering
  \caption{Rotation angle error for synthetic cryo-EM datasets (in degree).}
  \label{tab:2}
  \begin{tabular}{cccc}
    \hline\noalign{\smallskip}
                  & GroEL      & EMPIAR-10034 & EMPIAR-10065 \\
    \noalign{\smallskip}
    \hline
    \noalign{\smallskip}
    Fourier-based & 1.75       & 6.12         & 2.99         \\
    UDL-based     & {\bf 0.84} & {\bf 3.27}   & {\bf 2.33}   \\
    \hline
  \end{tabular}
\end{table}
\setlength{\tabcolsep}{1.4pt}

\begin{figure}
  \centering
  \includegraphics[height=2cm]{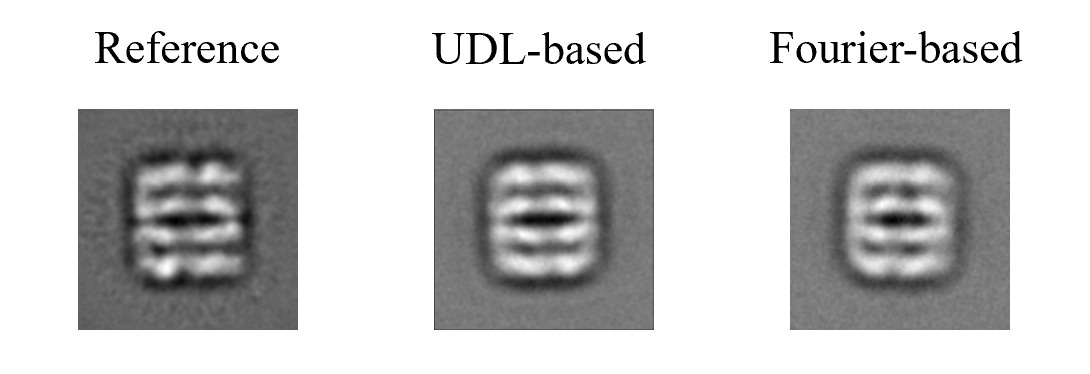}
  \caption{From left to right in the figure: reference image, average aligned image of our UDL-based method, average aligned image of traditional Fourier-based method.}
  \label{fig:6}
\end{figure}

As can be seen, we achieve more accurate rotation estimation than traditional Fourier-based method on synthetic cryo-EM datasets. For the real GroEL dataset, due to it only has 4694 images which is not enough for network training, we pre-train the network on synthetic GroEL dataset for 20 epochs and then initialize the network with pre-trained model to train on the real-world GroEL dataset. Since real GroEL dataset has no GTs, we align every image to a given reference image and then calculate their similarity using correntropy~\cite{ref035}. We also generate the average image of aligned images as shown in Fig.~\ref{fig:6}. The average similarity of aligned dataset is used to evaluate the performance of method, with ours' being 0.965 and the traditional one's being 0.961. Our proposed UDL-based unsupervised method achieves better results than traditional Fourier-based methods on both synthetic and real-world cryo-EM single particle datasets.

\section{Discussions}\label{sec:5}

From the perspective of equation solving, unknown variables (network outputs) and equations (loss functions using GTs) are equal in supervised learning problems. Whereas in unsupervised learning problems, naturally no equation is provided but unknown variables are required to be solved. Therefore, various unsupervised learning methods are proposed to create equations utilizing properties of specific tasks~\cite{ref005,ref039}. However, such specifically designed unsupervised strategy would be hard to generalize or transfer to other tasks.

In this paper, we propose a new difference learning strategy UDL which converts unsupervised learning to pseudo supervised learning. Though proposed to solve rigid image alignment in noisy cases, UDL essentially exploits the quantitative properties of network outputs in regression problem and thus could be generalized to certain regression problems. Unlike unsupervised learning methods designed for specific tasks which could create equal or even more equations than unknown variables (like pixel comparations in STN-based image alignment), UDL converts original problem containing no equation but unknown variables to problem containing equations and more unknown variables. In this way, though UDL essentially could not guarantee the convergence of training, it provides possible solutions containing the correct one. Then correct solutions could be obtained by adding restrictions or formulizing the difference functions, which depends on specific tasks. As discussed in Sect.~\ref{sec:3.2}, we use rotation angle as network output in rigid image alignment so that all solutions could be described by the correct solution plus a fixed bias. Though we cannot indicate a suitable difference function for each regression task due to lack of domain-specific knowledge, we hope that the UDL strategy could serve as a feasible method to help establish unsupervised methods for more regression tasks.

\section{Conclusions}

In this paper, we have proposed a new unsupervised difference learning strategy UDL for certain regression tasks and developed an unsupervised rigid image alignment method aimed at noisy cases based on it. UDL presents a new unsupervised learning strategy to convert unsupervised regression problem to pseudo supervised problem and its convergence can be achieved through designing difference functions based on specific tasks. It is expected that the new UDL pipeline will be extended to more applications in regression tasks.

\bibliographystyle{splncs04}
\bibliography{arXiv}

\begin{thebibliography}{10}
\providecommand{\url}[1]{\texttt{#1}}
\providecommand{\urlprefix}{URL }
\providecommand{\doi}[1]{https://doi.org/#1}

\bibitem{ref015}
Barnea, D.I., Silverman, H.F.: A class of algorithms for fast digital image
  registration. IEEE transactions on Computers  \textbf{100}(2),  179--186
  (1972). \doi{10.1109/TC.1972.5008923}

\bibitem{ref002}
Bendory, T., Bartesaghi, A., Singer, A.: Single-particle cryo-electron
  microscopy: Mathematical theory, computational challenges, and opportunities.
  IEEE signal processing magazine  \textbf{37}(2),  58--76 (2020).
  \doi{10.1109/MSP.2019.2957822}

\bibitem{ref030}
Bhamre, T., Zhang, T., Singer, A.: Denoising and covariance estimation of
  single particle cryo-em images. Journal of structural biology
  \textbf{195}(1),  72--81 (2016). \doi{10.1016/j.jsb.2016.04.013}

\bibitem{ref027}
Chen, J., He, Y., Frey, E.C., Li, Y., Du, Y.: Vit-v-net: Vision transformer for
  unsupervised volumetric medical image registration. arXiv preprint
  arXiv:2104.06468  (2021)

\bibitem{ref016}
Chen, Q.s., Defrise, M., Deconinck, F.: Symmetric phase-only matched filtering
  of fourier-mellin transforms for image registration and recognition. IEEE
  Transactions on pattern analysis and machine intelligence  \textbf{16}(12),
  1156--1168 (1994). \doi{10.1109/34.387491}

\bibitem{ref019}
Chen, Y.X., Xie, R., Yang, Y., He, L., Feng, D., Shen, H.B.: Fast cryo-em image
  alignment algorithm using power spectrum features. Journal of Chemical
  Information and Modeling  \textbf{61}(9),  4795--4806 (2021).
  \doi{10.1021/acs.jcim.1c00745}

\bibitem{ref004}
DeTone, D., Malisiewicz, T., Rabinovich, A.: Deep image homography estimation.
  arXiv preprint arXiv:1606.03798  (2016)

\bibitem{ref026}
Dosovitskiy, A., Beyer, L., Kolesnikov, A., Weissenborn, D., Zhai, X.,
  Unterthiner, T., Dehghani, M., Minderer, M., Heigold, G., Gelly, S., et~al.:
  An image is worth 16x16 words: Transformers for image recognition at scale.
  arXiv preprint arXiv:2010.11929  (2020)

\bibitem{ref014}
Fischler, M.A., Bolles, R.C.: Random sample consensus: a paradigm for model
  fitting with applications to image analysis and automated cartography.
  Communications of the ACM  \textbf{24}(6),  381--395 (1981).
  \doi{10.1145/358669.358692}

\bibitem{ref018}
Frank, J., Radermacher, M., Penczek, P., Zhu, J., Li, Y., Ladjadj, M., Leith,
  A.: Spider and web: processing and visualization of images in 3d electron
  microscopy and related fields. Journal of structural biology
  \textbf{116}(1),  190--199 (1996). \doi{10.1006/jsbi.1996.0030}

\bibitem{ref024}
Hartley, R., Zisserman, A.: Multiple view geometry in computer vision.
  Cambridge university press (2003)

\bibitem{ref028}
He, K., Zhang, X., Ren, S., Sun, J.: Deep residual learning for image
  recognition. In: Proceedings of the IEEE conference on computer vision and
  pattern recognition. pp. 770--778 (2016)

\bibitem{ref037}
Ioffe, S., Szegedy, C.: Batch normalization: Accelerating deep network training
  by reducing internal covariate shift. In: International conference on machine
  learning. pp. 448--456. PMLR (2015)

\bibitem{ref033}
Iudin, A., Korir, P.K., Salavert-Torres, J., Kleywegt, G.J., Patwardhan, A.:
  Empiar: a public archive for raw electron microscopy image data. Nature
  methods  \textbf{13}(5),  387--388 (2016). \doi{10.1038/nmeth.3806}

\bibitem{ref006}
Jaderberg, M., Simonyan, K., Zisserman, A., et~al.: Spatial transformer
  networks. Advances in neural information processing systems  \textbf{28}
  (2015)

\bibitem{ref031}
Joyeux, L., Penczek, P.A.: Efficiency of 2d alignment methods. Ultramicroscopy
  \textbf{92}(2),  33--46 (2002). \doi{10.1016/S0304-3991(01)00154-1}

\bibitem{ref036}
Kingma, D.P., Ba, J.: Adam: A method for stochastic optimization. arXiv
  preprint arXiv:1412.6980  (2014)

\bibitem{ref008}
Koguciuk, D., Arani, E., Zonooz, B.: Perceptual loss for robust unsupervised
  homography estimation. In: Proceedings of the IEEE/CVF Conference on Computer
  Vision and Pattern Recognition. pp. 4274--4283 (2021)

\bibitem{ref039}
Krull, A., Buchholz, T.O., Jug, F.: Noise2void-learning denoising from single
  noisy images. In: Proceedings of the IEEE/CVF Conference on Computer Vision
  and Pattern Recognition. pp. 2129--2137 (2019)

\bibitem{ref010}
Lin, T.Y., Maire, M., Belongie, S., Hays, J., Perona, P., Ramanan, D.,
  Doll{\'a}r, P., Zitnick, C.L.: Microsoft coco: Common objects in context. In:
  European conference on computer vision. pp. 740--755. Springer (2014).
  \doi{10.1007/978-3-319-10602-1\_48}

\bibitem{ref012}
Lowe, D.G.: Distinctive image features from scale-invariant keypoints.
  International journal of computer vision  \textbf{60}(2),  91--110 (2004).
  \doi{10.1023/B:VISI.0000029664.99615.94}

\bibitem{ref034}
Ludtke, S.J., Chen, D.H., Song, J.L., Chuang, D.T., Chiu, W.: Seeing groel at 6
  {\aa} resolution by single particle electron cryomicroscopy. Structure
  \textbf{12}(7),  1129--1136 (2004). \doi{10.1016/j.str.2004.05.006}

\bibitem{ref001}
Monnier, T., Groueix, T., Aubry, M.: Deep transformation-invariant clustering.
  Advances in Neural Information Processing Systems  \textbf{33},  7945--7955
  (2020)

\bibitem{ref005}
Nguyen, T., Chen, S.W., Shivakumar, S.S., Taylor, C.J., Kumar, V.: Unsupervised
  deep homography: A fast and robust homography estimation model. IEEE Robotics
  and Automation Letters  \textbf{3}(3),  2346--2353 (2018).
  \doi{10.1109/LRA.2018.2809549}

\bibitem{ref038}
Paszke, A., Gross, S., Massa, F., Lerer, A., Bradbury, J., Chanan, G., Killeen,
  T., Lin, Z., Gimelshein, N., Antiga, L., et~al.: Pytorch: An imperative
  style, high-performance deep learning library. Advances in neural information
  processing systems  \textbf{32} (2019)

\bibitem{ref032}
Punjani, A., Zhang, H., Fleet, D.J.: Non-uniform refinement: adaptive
  regularization improves single-particle cryo-em reconstruction. Nature
  methods  \textbf{17}(12),  1214--1221 (2020).
  \doi{10.1038/s41592-020-00990-8}

\bibitem{ref021}
Rocco, I., Arandjelovic, R., Sivic, J.: Convolutional neural network
  architecture for geometric matching. In: Proceedings of the IEEE conference
  on computer vision and pattern recognition. pp. 6148--6157 (2017)

\bibitem{ref023}
Ronneberger, O., Fischer, P., Brox, T.: U-net: Convolutional networks for
  biomedical image segmentation. In: International Conference on Medical image
  computing and computer-assisted intervention. pp. 234--241. Springer (2015).
  \doi{10.1007/978-3-319-24574-4\_28}

\bibitem{ref017}
De~la Rosa-Trev{\'\i}n, J., Ot{\'o}n, J., Marabini, R., Zald{\'\i}var, A.,
  Vargas, J., Carazo, J., Sorzano, C.: Xmipp 3.0: an improved software suite
  for image processing in electron microscopy. Journal of structural biology
  \textbf{184}(2),  321--328 (2013). \doi{10.1016/j.jsb.2013.09.015}

\bibitem{ref003}
Rubinstein, M., Joulin, A., Kopf, J., Liu, C.: Unsupervised joint object
  discovery and segmentation in internet images. In: Proceedings of the IEEE
  conference on computer vision and pattern recognition. pp. 1939--1946 (2013)

\bibitem{ref013}
Rublee, E., Rabaud, V., Konolige, K., Bradski, G.: Orb: An efficient
  alternative to sift or surf. In: 2011 International conference on computer
  vision. pp. 2564--2571. Ieee (2011). \doi{10.1109/ICCV.2011.6126544}

\bibitem{ref020}
Simonyan, K., Zisserman, A.: Very deep convolutional networks for large-scale
  image recognition. arXiv preprint arXiv:1409.1556  (2014)

\bibitem{ref009}
Sloan, J.M., Goatman, K.A., Siebert, J.P.: Learning rigid image
  registration-utilizing convolutional neural networks for medical image
  registration  (2018). \doi{10.5220/0006543700890099}

\bibitem{ref035}
Sorzano, C.O.S., Bilbao-Castro, J., Shkolnisky, Y., Alcorlo, M., Melero, R.,
  Caffarena-Fern{\'a}ndez, G., Li, M., Xu, G., Marabini, R., Carazo, J.: A
  clustering approach to multireference alignment of single-particle
  projections in electron microscopy. Journal of structural biology
  \textbf{171}(2),  197--206 (2010). \doi{10.1016/j.jsb.2010.03.011}

\bibitem{ref025}
Vaswani, A., Shazeer, N., Parmar, N., Uszkoreit, J., Jones, L., Gomez, A.N.,
  Kaiser, {\L}., Polosukhin, I.: Attention is all you need. Advances in neural
  information processing systems  \textbf{30} (2017)

\bibitem{ref022}
Zeng, R., Denman, S., Sridharan, S., Fookes, C.: Rethinking planar homography
  estimation using perspective fields. In: Asian Conference on Computer Vision.
  pp. 571--586. Springer (2018). \doi{10.1007/978-3-030-20876-9\_36}

\bibitem{ref029}
Zeng, X., Xu, M.: Gum-net: Unsupervised geometric matching for fast and
  accurate 3d subtomogram image alignment and averaging. In: Proceedings of the
  IEEE/CVF conference on computer vision and pattern recognition. pp.
  4073--4084 (2020)

\bibitem{ref007}
Zhang, J., Wang, C., Liu, S., Jia, L., Ye, N., Wang, J., Zhou, J., Sun, J.:
  Content-aware unsupervised deep homography estimation. In: European
  Conference on Computer Vision. pp. 653--669. Springer (2020).
  \doi{10.1007/978-3-030-58452-8\_38}

\bibitem{ref011}
Zitova, B., Flusser, J.: Image registration methods: a survey. Image and vision
  computing  \textbf{21}(11),  977--1000 (2003).
  \doi{10.1016/S0262-8856(03)00137-9}

\end{thebibliography}
\end{document}